%% file: submission.tex
\documentclass[10pt,letterpaper]{article}
\usepackage[top=0.85in,left=1.2in,footskip=0.75in,marginparwidth=5.5in]{geometry}

\usepackage[utf8]{inputenc}

\usepackage{cite}
\usepackage{amssymb}
\usepackage{amsmath}
\usepackage{nameref,hyperref}

\usepackage[right]{lineno}

\usepackage{microtype}
\DisableLigatures[f]{encoding = *, family = * }

\usepackage{changepage}

\usepackage[aboveskip=1pt,labelfont=bf,labelsep=period,singlelinecheck=off]{caption}

\makeatletter
\renewcommand{\@biblabel}[1]{\quad#1.}
\makeatother

\usepackage{lastpage,fancyhdr,graphicx}
\usepackage{epstopdf}
\usepackage{color}

\definecolor{Gray}{gray}{.25}

\usepackage{graphicx}

\usepackage{sidecap}

\usepackage{wrapfig}
\usepackage[pscoord]{eso-pic}
\usepackage[fulladjust]{marginnote}
\reversemarginpar

\begin{document}
\vspace*{0.35in}

\begin{flushleft}
{\Large
\textbf\newline{Global Consistent Point Cloud Registration Based on Lie-algebraic Cohomology }
}
\newline
\\
Yuxue Ren\textsuperscript{1},
Baowei Jiang\textsuperscript{2},
Wei Chen\textsuperscript{3},
Na Lei\textsuperscript{3,*},
Xianfeng David Gu\textsuperscript{4},
\\
\bigskip
\bf{1}  Academy for Multidisciplinary Studies, Capital Normal University
\\
\bf{2} Beijing Advanced Innovation Center for Imaging Theory and Technology, Capital Normal University
\\
\bf{3} DUT-RU ISE, Dalian University of Technology
\\
\bf{4} State University of New York at Stony Brook
\\
\bigskip
* correseponding nalei@dlut.edu.cn

\end{flushleft}

\input{abstract}

\input{introduction}
\input{previous}

\input{theory}

\input{algorithm}
\input{experiments}
\input{conclusion}

\bibliography{egbib}

\bibliographystyle{abbrv}

\end{document}

%% file: abstract.tex
 \section*{Abstract}
We present a novel, effective method for global point cloud registration problems by geometric topology. Based on many point cloud pairwise registration methods (e.g ICP), we focus on the problem of accumulated error for the composition of transformations along any loops.
The major technical contribution of this paper is a linear method for the elimination of errors, using only solving a Poisson equation. We demonstrate the consistency of our method from Hodge-Helmhotz decomposition theorem and experiments on multiple RGBD datasets of real-world scenes. The experimental results also demonstrate that our global registration method runs quickly and provides accurate reconstructions.

\textbf{keywords} Global Point Cloud Registration, Lie-algebraic cohomology, Hodge-Helmhotz decomposition, graph theory

%% file: introduction.tex
\section{Introduction}
\label{sec:intro}



\noindent{\bf Motivation} Recent years have witnessed the rapid development of Structure from Motion (SfM)\cite{5995626,5539801,ozyesil2017survey,7410455}  and Simultaneous Localization And Mapping (SLAM) \cite{6907055,6906953,Strasdat,ELCH}. Many global reconstruction methods have been proposed, among them the most traditional ones are the sequential approaches, which iteratively add new data to 3D models from point cloud collections. In practice, sequential approaches are expensive, because they require repeated nonlinear model refinement (bundle adjustment)~\cite{7780814,BundleFusion}; they are error-proning, since they may lead to error accumulation and exacerbate the drifting effects. 

To improve the consistency, 
the fundamental concept of \emph{viewing graph} has been developed. The viewing graph encapsulates the point clouds treated as nodes and the relative transformations between the point clouds as edges. 
\emph{Global consistency} is defined as concatenating transformations along each loop in the graph. It should return the identity under the assumption of an ideal noise-free setting. 

Based on the viewing graph, many iterative reconstruction methods have been proposed~\cite{lim_online_2011,4587678,5995626,5539801,Zhou2013}. These methods consider all relative poses of point clouds (e.g. edges of the graph), simultaneously estimate all poses in a single step and restrict the focus on cycles in the graph structure~\cite{6906953,1315098,ELCH,5539801}. Although sophisticated techniques have been applied, such as nonlinear optimization~\cite{LUM,ELCH} and ~\cite{1307010,1315098,5995745,arrigoni2020synchronization} rotation averaging, the existing methods cannot achieve global consistency with high precision, only obtain rough approximated solutions. Our goal is to propose a novel algorithm to tackle the loop close problem and greatly improve the global consistency.\\

\noindent{\bf Our Solution} In this work, we approach the global consistency problem from a fundamentally different perspective: \emph{the viewing graph is treated not as a solely 1-skeleton graph, but a higher dimensional Cech complex, namely it has much richer topological information, encoded by the Cech cohomology, which is the key to solve the loop closing problem}. \\
\noindent{\bf Contribution} This work proposes a novel algorithm to tackle the loop closing problem based on Lie-Algebraic Cech cohomology, the contributions include:\\
\noindent{1. Our method is different with other graph optimization method, based on Helmholtz-Hodge decomposition theorem, the loop closing problem can be solved through solving a system of linear equations.} \\
\noindent{2. The viewing graph is generalized to a Cech complex to enrich the topological structure.}\\
\noindent{3. The relative rigid motion on the edges is formulated as a Lie-algebraic simplicial 1-form, thus the loop closing condition is equivalent to the exactness of the 1-form.}\\
\noindent{4. The exact component of the Lie-algebraic 1-form can be extracted using Helmholtz-Hodge decomposition algorithm.}\\
\noindent{5. The 2-skeleton of the Cech complex can be approximated by graph embedding.}\\
Furthermore, our experimental results demonstrate the simplicity, efficiency and accuracy of the proposed algorithm.

The remainder of the paper is organized as follows: 
the related works are reviewed in Sec. 2; the elementary concepts are introduced in Sec. 3; the computational algorithms are explained in Sec. 4, including Cech complex construction, graph embedding, Lie-algebra valued 1-form and Hodge decomposition; the experimental results are reported in Sec. 5; and eventually the paper is concluded in Sec. 6.

%% file: previous.tex
\section{Related Work}
In this section, we review the most related approaches for solving the loop closing problem. There are mainly three types of algorithms: matrix methods, averaging rotation and translation, graph optimization.\\
\noindent{\textbf{Matrix Methods}} Arie-Nachimson~\cite{6374980} constructed a $3n\times 3n$ symmetric matrix by concatenating the pairwise rotation matrices and then used either the spectral decomposition or the semidefinite programming (SDP) method to compute the set of global rotations. Arrigoni~\cite{7035862} also views the absolute rotation estimation problem as a low-rank and sparse matrix decomposition for the above symmetric matrix by Bilateral Random Projections (BRP). They calculated the rotation and the translation separately. In contrast, our method computes the rotation and the translation simultaneously.\\
\noindent{\textbf{Averaging Rotation}} Govindu~\cite{1315098,Govindu2006,govindu2013averaging} and Chatterjee~\cite{6751174} used the lie-algebraic averaging method to control the drift due to the accumulating error and the estimated camera motion. Hartley~\cite{5995745,Hartley2012RotationA} gave a provably convergent algorithm for finding the $L^1$ mean of $SO(3)$ under several relative orientation measurements. The work in~\cite{1307010} proposed edge weights method to average rotation. All the above works decouple translations from rotations.
Purkait~\cite{purkait_neurora_2020} viewed relative orientations,i.e. rotation between two frames, as the edge features, then proposed the view-graph cleaning network (CleanNet) and fine-tuning network (FineNet), which built on Message-Passing Neural Networks (MPNN), to predict outlier edges and refined absolute orientations respectively. Taking the average can ensure consistency along special loops, but it is not clear whether the consistency holds for all the loops. The proposed Hodge decomposition method can guarantee global consistency along any loop of the viewing graph.\\
\noindent{\textbf{Graph Optimization}} Snavely~\cite{lim_online_2011} used a graph algorithm-based maximum leaf spanning tree (MLST) to select a skeletal subset of images and finally used bundle adjustment, which aimed at reducing the computation redundancy. And Lim~\cite{lim_online_2011} proposed a system consisting of a local as well as a global adjustment on the online environment mapping, which obtains a key-frame pose graph from a key-frames subset. Sprickerhof~\cite{ELCH} proposed a vertex weights algorithm to optimize the graph by removing edge. For the learning method, the set of edges with the bad relationship can be inferred by a Bayesian framework in~\cite{5539801}. This type of methods destroy the original viewing graph, therefore lose partial topological information of the Cech complex. Compared to the existing methods, our proposed method achieves high accuracy, and the optimized viewing graph does not require any filtering steps to remove “bad” relative poses ~\cite{4587678,lim_online_2011,5539801}. Hence, our algorithmic pipeline is simpler and more efficient than alternative approaches~\cite{ELCH,1315098,5995745}.

\noindent{\textbf{Helmholtz-Hodge Decomposition Theorem}}
The well-known Helmholtz-Hodge decomposition theorem ~\cite{lui2014shape,Desbrun,6365629} claims that any differential 1-form on a Riemannian manifold can be decomposed into three orthogonal components, an exact form (curl free), a co-exact form (divergence free) and a harmonic form (both curl free and divergence free). Namely, each cohomological class has a unique harmonic form. Therefore, we can extract the exact component from the initial Lie-algebraic 1-form.
Unlike existing methods~\cite{1315098,Govindu2006,Krishnan,Tomasi,4270140}, our algorithm mentioned in Section 4 solves global rotations and translations simultaneously linearly.\\

%% file: theory.tex
\section{Theoretic Background}
\subsection{Cech Complex and Cech Coholomopy} 
In the viewing graph, each point cloud is treated as a node, two overlapping point clouds are represented as an edge, three point clouds mutually intersecting each other should be represented as a 2-dimensional face (2-simplex) in the Cech complex, similarly each $k$-order cliques in the viewing graph is a $k$ dimensional simplex in the Cech complex. The Cech complex enriches the topological and the algebraic relation of the viewing graph~\cite{Bott}. In more detail, let $\mathfrak{U}=\{U_\alpha\}_{\alpha \in J}$ be a set of point clouds, where the index set $J$ is a countable ordered set. Every $U_{\alpha}$ can also be regarded as an open set embedded in 3D space from the geometric point of view. Denote the  intersection set $U_\alpha \bigcap U_\beta$ as $U_{\alpha\beta}$, etc. The definition of Cech complex is as follows:\\
\textbf{Definition 1 (Cech complex)}
The Cech complex of $\mathfrak{U}$ is a simplicial complex constructed as follows. To every open set $U_\alpha$, we associate a vertex $\alpha$. If $U_{\alpha_0} \bigcap U_{\alpha_1}$ is nonempty, we connect the vertices $\alpha_0$ and $\alpha_1$ with an edge, if $U_{\alpha_0}\cap U_{\alpha_1}\cap\cdots\cap U_{\alpha_n}$ is non empty, we say $\alpha_0,\alpha_1,\cdots,\alpha_n$ construct an $n$ dimensional cell of the Cech complex etc. For the basics of simplicial complexes, see\cite{croom2012basic}.
\begin{figure}
\centering
\includegraphics[width=0.45\textwidth]{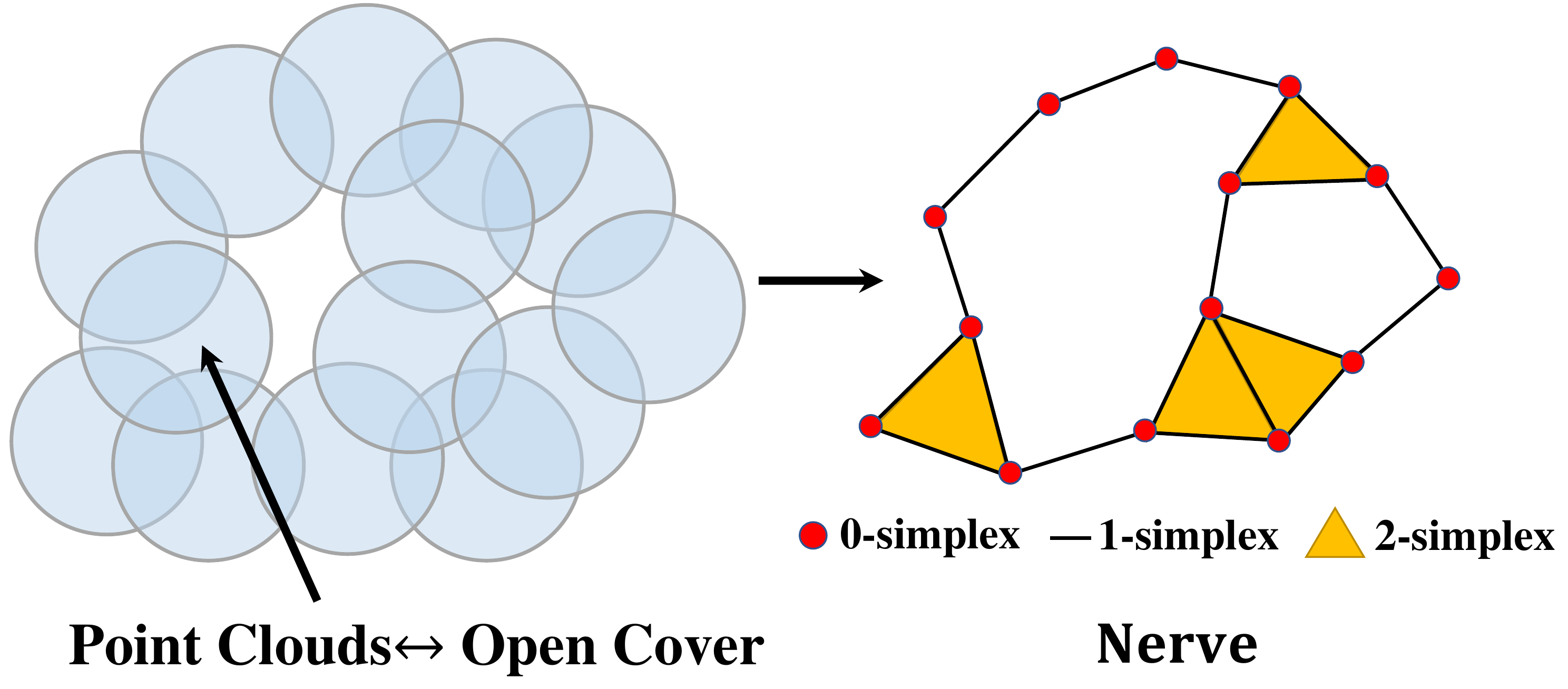}
\caption{Construction of Cech complex}
\label{Cech complex}
\end{figure}

A function defined on $k$-dimensional simplicial of Cech complex is represented as a $k$-form, all of this kind of functions form a topological space, i.e., Cech Coholomopy.

\textbf{Cech Coholomopy}
Let $\Omega^0(U)$ be the vector space of const functions on $U$, and $\Omega^1(U)$ be the space of 1-forms on $U$ ,etc. Let $C^0(\mathfrak{U},\mathbb{R})=\prod_{\alpha_0}\Omega^0(U_{\alpha_0})$ be the connection of all const functions on $U_\alpha$, and $C^1(\mathfrak{U},\mathbb{R})=\prod_{\alpha_0\alpha_1}\Omega^0(U_{\alpha_0\alpha_1})$ be the connection of const functions on all $U_{\alpha_0\alpha_1}$. Following a similar approach,  $C^{n+1}(\mathfrak{U},\mathbb{R})=\prod_{\alpha_0\cdots\alpha_{n+1}}\Omega^0(U_{\alpha_0\cdots\alpha_{n+1}})$ can be defined the connection of const functions on all $U_{\alpha_0\alpha_1\cdots\alpha_{n+1}}$. For $\omega\in C^p(\mathfrak{U},\mathbb{R})$, and $\partial_q:C^{p}(\mathfrak{U},\mathbb{R})\to C^{p+1}(\mathfrak{U},\mathbb{R})$, define the boundary operator as
\begin{equation} (\partial_q\omega)_{\alpha_0\cdots\alpha_{p+1}}=\sum_{i=0}^{p+1}(-1)^i\omega_{\alpha_0\cdots\hat{\alpha_i}\cdots\alpha_{p+1}} .
\end{equation}

\textbf{Definition 2 (Cocycle)} A cochain $\omega\in C^q(\mathfrak{U},\mathbb{R})$ is the $\partial_q$-cocycle that satisfies $\partial_q \omega = 0$, namely the kernel of $\partial_q$.

\textbf{Definition 3 (Coboundary)} A cochain $\omega\in C^q(\mathfrak{U},\mathbb{R})$ is called $\partial_q$-coboundary, if there exists a $\omega^{'} \in C^{q-1}(\mathfrak{U},\mathbb{R})$ satisfies $\partial_{q-1} \omega^{'} = \omega$, namely the image of $\partial_{q-1}$.

In this problem, the Lie-algebraic of the relative rigid motion on the edges of Cech complex is formulated as a simplicial 1-form, and loop closing condition corresponds to an exact 1-form, i.e., a 1-form satisfying cocycle condition.
\subsection{Cocycle Condition}
Let $\{\omega_{\alpha\beta}:U_{\alpha\beta}\rightarrow\mathbb{R},\forall U_{\alpha\beta}\}$ be a 1-form defined on edges of Cech complex, for the transition function $\{\omega_{\alpha\beta}\}$ of the open cover $\mathfrak{U}$, if they satisfy
\begin{equation}
	\omega_{\alpha\beta}\circ \omega_{\beta\gamma} = \omega_{\alpha\gamma} \quad \text{on} \quad U_\alpha \cap U_\beta\cap U_\gamma ,
\end{equation}
for any open subsets $U_\alpha,U_\beta,U_\gamma \in \mathfrak{U}$, then $\{\omega_{\alpha\beta}\}$ satisfies the co-cycle condition. In more general cases, for $U_{\alpha_i}$, where $\alpha_i \in J, i=1,\cdots,n$, co-cycle condition is 
\begin{equation}
	\omega_{\alpha_0\alpha_1}\omega_{\alpha_1\alpha_2}\cdots\omega_{\alpha_{n-1}\alpha_{n}}=\omega_{\alpha_0\alpha_{n}} .
\end{equation}

When the set of point clouds are registered together, the error accumulation of local registration between point clouds leads to large errors for the first and last point clouds in a closed loop. This error is essentially the product of transformations on the closed loop, not the identity mapping. The condition requires that the product of transformation on any closed loop is identical mapping, so that any closed loop has global consistency.

From the view of the Cech cohomology above, coboundaries, satisfy the cocycle condition naturally, but the cocycles may not. In other words, if a transition function $\{\omega_{ij}\}$ of $\mathfrak{U}$ is a coboundary, then the product on every closed loop is identity.

The next section introduces the way to compute a coboundary from a cocycle.
\subsection{Lie Algebras of the relative rigid motion}
All the rigid motions form a Lie group, the tangent space at the identity form the Lie algebra. Since the rotation matrix is not commutative, during the process of our algorithm, the Lie algebras of the Rotation and Euclidean groups should be used. For rotation $R\in SO(3)$, its lie-algebra $\phi\in \mathbb{R}$ satisfy $R=exp(\phi^{\wedge})$, where
\begin{equation}
    \phi^{\wedge}=\left[
    \begin{array}{ccc}
    0 & -\phi_{3}  & \phi_{2} \\
    \phi_{3} & 0  & -\phi_{1} \\
     -\phi_{2} & \phi_{1} & 0 \\
    \end{array}
    \right] .
\end{equation} 
Then for rotation $R\in SO(3)$ and translation $t\in \mathbb{R}^{3}$, their lie-algebra $[\phi,t] \in \mathbb{R}^{6}$, is Abelian. Thus the 6D vector on the graph is our motion field. For each oriented edge of the complex, we associate a rigid motion with it, which can be estimated by the iterative closest point (ICP) algorithm. Then we further represent the rigid motion as a vector in the Lie algebra. Therefore we can treat the rigid motions defined on all the edges as a Lie-algebraic simplicial 1-form on the Cech complex, then use the Lie-algebraic simplicial cohomology to rephrase the loop close problem: the integration of the Lie-algebraic 1-form is zero along any closed loop (1-chain), this implies the 1-form is exact, namely it is the gradient of some function defined on the Cech complex.

\subsection{Graph Embedding}
The topological structure of the Cech complex is complicated, involving high dimensional simplexes. For Hodge decomposition, only 2-skeleton is needed. Therefore, we can embed the view graph on a surface, and use the surface to approximate the 2-skeleton of the Cech complex and perform the Hodge decomposition. This method preserves the algebraic property of the viewing graph~\cite{Lee}, ensures the loop closing condition and improves the efficiency.

An embedding of $G$ is a map $f:G\to S$, such that each edge $e\in E_G$ is mapped to a simple arc on the surface, and these arcs don't intersect at the interior points~\cite{gross2001topological,lando2004graphs}. It is well known that embedding a graph onto a surface with a fixed genus is linear time in the graph size and doubly exponential in the genus~\cite{mohar1999linear}, while finding the embedding surface with the minimal genus is NP-hard ~\cite{thomassen1989graph}. For the current purpose, we choose a linear algorithm and do not pay attention to minimizing the genus.

%% file: algorithm.tex
\section{Computational Algorithms}
\label{sec:algorithm}
\begin{figure} 
\centering 
\includegraphics[width=1.0\linewidth]{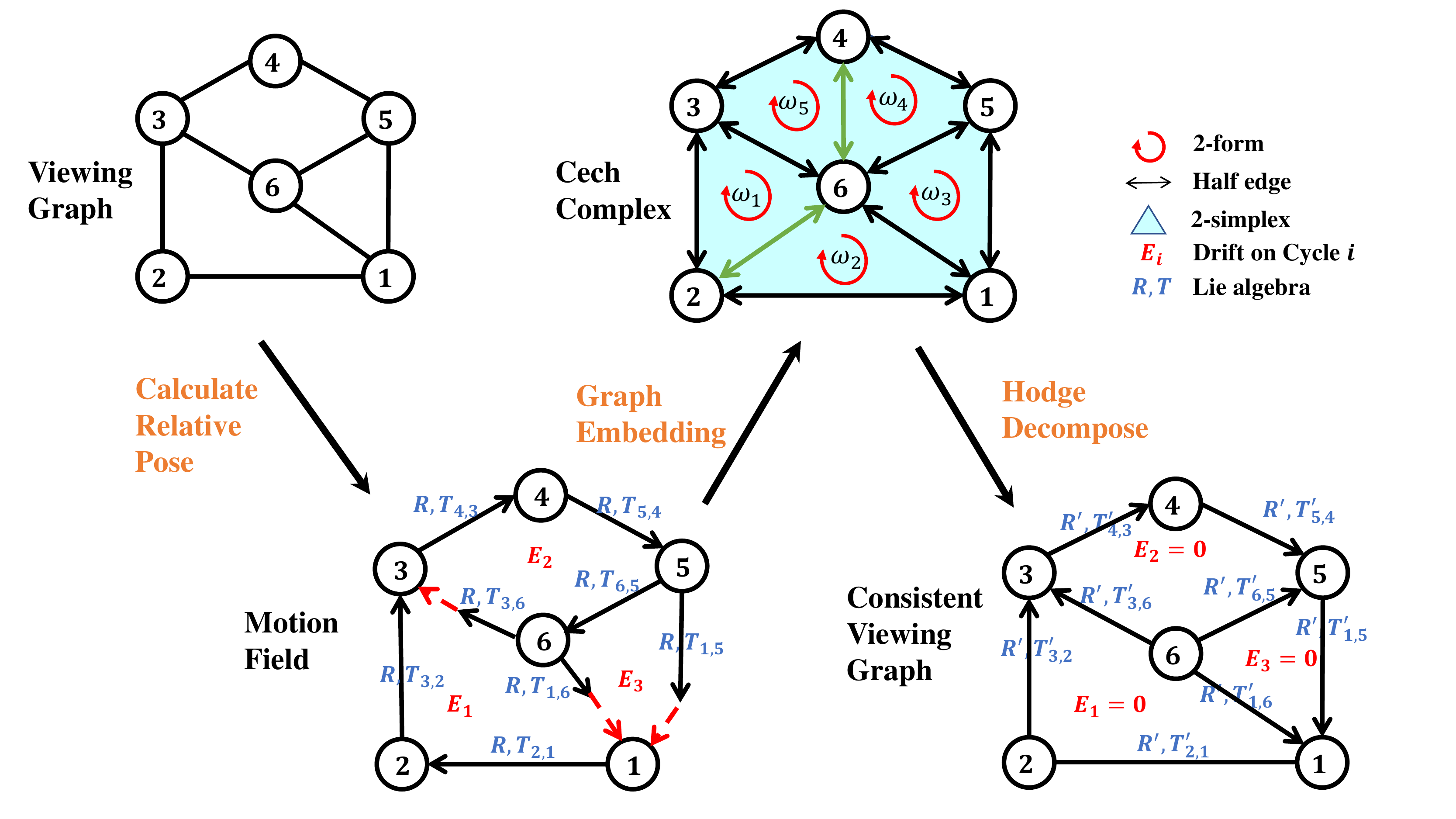}
\caption{Hodge Algorithms Pipeline. a) Construct viewing graph by SIFT (if the number of matches is more than 50, add the corresponding edge), b) calculate Lie algebra  of rigid transform from ICP, c) triangularization by graph embedding, 4) solve curl free field by solving equation \ref{eqn:discrete_Poisson}. Finally, we realize consistency.}
\label{Pipeline}
\end{figure}

In this section, we explain our algorithmic pipeline in detail, Fig~\ref{Pipeline} lists the pipline.\\
\noindent{\bf Cech Complex Construction}
The viewing graph $G=(V_G,E_G)$ is constructed as usual, each node in $V_G$ represents a point cloud, the corresponding camera frame is called a pose. The point clouds are roughly registered using ICP or NDT algorithm. Each edge in $E_G$ represents two intersecting point clouds, and is associated with the transformation between the two poses associated with the end nodes. The viewing graph is the 1-skeleton of the Cech complex. Each $k$-clique ($k$-complete sub-graph) of the viewing graph represents a $k$-dimensional simplex in the Cech complex, its boundary consists of all the $k-1$ sub-cliques. The Cech complex is denoted as $\Sigma$. \\

\noindent{\bf Graph Embedding} The Cech complex $\Sigma$ has a complicated topological structure, we can embed the viewing graph onto a surface and use the embedding surface to approximate the 2-skeleton of $\Sigma$. 
For each vertex $v\in V_G$, we randomly define a cyclic order for all the edges adjacent to $v$, denoted as $e_0,e_1,\dots,e_{n-1}$. For each edge we define two half-edges $h_k^+$ and $h_k^-$, where the target vertex of $h_k^+$ is $v$ and the source vertex of $h_k^-$ is $v$. After defining the cyclic orders for all the vertices and constructing all the halfedges, we connect the halfedges as follows: for each vertex $v$ and for each $k$, we connect $h_k^+$ to $h_{k+1}^-$, where $k+1$ is modulo $n$. After connecting all the halfedges, if we start an arbitrary halfedge, and trace it by the connection, we will go through a loop and return to the initial halfedge. Namely, we have divided all the halfedges into disjoint loops, each loop is a 2-dimensional topological cell. All the cells are glued together by the edges to form an oriented closed surface. This linear algorithm gives one embedding surface of the graph $G$, denoted as $S$. Furthermore, we can triangulate each cell and convert $S$ and obtain a topological triangulation of $S$. We represent $S$ as a triangle mesh. The details can be found in Alg.1.\\

\setlength{\tabcolsep}{4pt}
\begin{table}
\begin{center}
\label{alg:graph_embedding}
\begin{tabular}{l}
\hline\noalign{\smallskip}
{\bf Algorithm 1} Graph Embedding\\
\noalign{\smallskip}
\hline
\noalign{\smallskip}
{\bf Input:} A 3-connected Graph $G=(V_G,E_G)$\\
{\bf Output:} A surface $S$ and an embedding $f:G\to S$\\ \hline
{\bf For} each vertex $v$ \\
\quad Define a cyclic order of edges adjacent to $v$; \\
\quad Attach two opposite halfedges $h_k^+,h_k^-$ to each edge $e_k$; \\
\quad The target vertex of $h_k^+$ is $v$; Connect $h_k^{+}$ to $h_{k+1}^-$;\\
{\bf End For}\\
{\bf For All} halfedge $h$\\
\quad Trace $h$ to form a loop;\\
{\bf End For}\\
Each loop gives a cell; All the cells are glued by the edges to form a surface $S$;\\
Triangulate the surface $S$.\\
\hline
\end{tabular}
\end{center}
\end{table}
\setlength{\tabcolsep}{1.4pt}

\noindent{\bf Rigid Transformation 1-Form}
Each edge of the viewing graph is associated with a relative point cloud transforming the point clouds associated with the two end nodes, which are obtained by point cloud registration. The transformations on the edges don't satisfy the loop closing condition, we have to compose the transformation on each edge with a small adjustment to meet the constraint. The small adjustment for the relative transformation on each edge is a differential rigid motion, denoted as $\omega=(\omega_1,\omega_2)$, where $\omega_1: E_G\to \mathfrak{so}(3)$ is the rotation component taking the value in the Lie-algebra $\mathfrak{so}(3)$, $\omega_2:E_G\to \mathbb{R}^3$ the translation component. $\omega$ is treated as a Lie-algebraic 1-form defined on $\Sigma$. Consider each edge $e$ in the triangle mesh $S$, if $e$ is in the viewing graph $G$, then $\omega(e)$ has been calculated already. If $e$ is not in the graph $G$, then on the graph $G$ there is the shortest path $\gamma$ connecting the two vertices of $e$, therefore, we define $\omega(e)$ equals to the integration of $\omega$ along $\gamma$, namely
\begin{equation}
    \omega(e) = \int_\gamma \omega = \sum_{e_i\in \gamma \subset E_G} \omega(e_i) .
\end{equation}
The details of the algorithm can be found in Alg.2.

\setlength{\tabcolsep}{4pt}
\begin{table}
\begin{center}
\label{alg:transformation_form}
\begin{tabular}{l}
\hline\noalign{\smallskip}
{\bf Algorithm 2} Rigid Transformation Form\\
\noalign{\smallskip}
\hline
\noalign{\smallskip}
{\bf Input:} The viewing graph $G=(V_G,E_G)$; \\
the embedding surface $S$; \\
the rigid motions defined on $G$, $\omega_0$.\\
{\bf Output:} The rigid transformation form $\omega$ defined on $S$.\\ 
\hline
{\bf For} each edge $e$ in S\\
\quad{\bf If} $e$ is not in $G$\\ 
\qquad Find the shortest path $\gamma\subset G$ connecting the two end vertices of $e$;\\
\qquad Define $\omega(e)$ as the integration of $\omega_0$ along $\gamma$;\\
\quad{\bf End If}\\
{\bf End For}\\
\hline
\end{tabular}
\end{center}
\end{table}
\setlength{\tabcolsep}{1.4pt}


\noindent{\bf Hodge Decomposition}\\
\textbf{Theorem 4 (Hodge Decomposition Theorem)}\quad For any $\omega\in C^q(\mathfrak{U},\mathbb{R})$, there is a unique decomposition as
\begin{equation}
    \omega=df+\delta g+h .
\label{eqn:harmonic}
\end{equation}
where $df$ is an exact form, $\delta g$ is a co-exact form and $h$ is a harmonic form.

According to Helmholtz-Hodge theorem, $\omega=(\omega_1,\omega_2)$ can be decomposed into the sum of an exact form, a co-exact form and a harmonic form, and the decomposition is unique. 
The work of ~\cite{lui2014shape} gives an algorithm to extract the harmonic component from a given 1-form. In this work, we apply the similar method, from Eqn.~\ref{eqn:harmonic}, we have
\begin{equation}
\delta\omega = \delta df + \delta^2 g + \delta h = \Delta f ,     \label{eqn:Poisson}
\end{equation}
therefore the function $f$ satisfies the Poisson equation.

\begin{figure}
\centering
\includegraphics[width=0.2\textwidth]{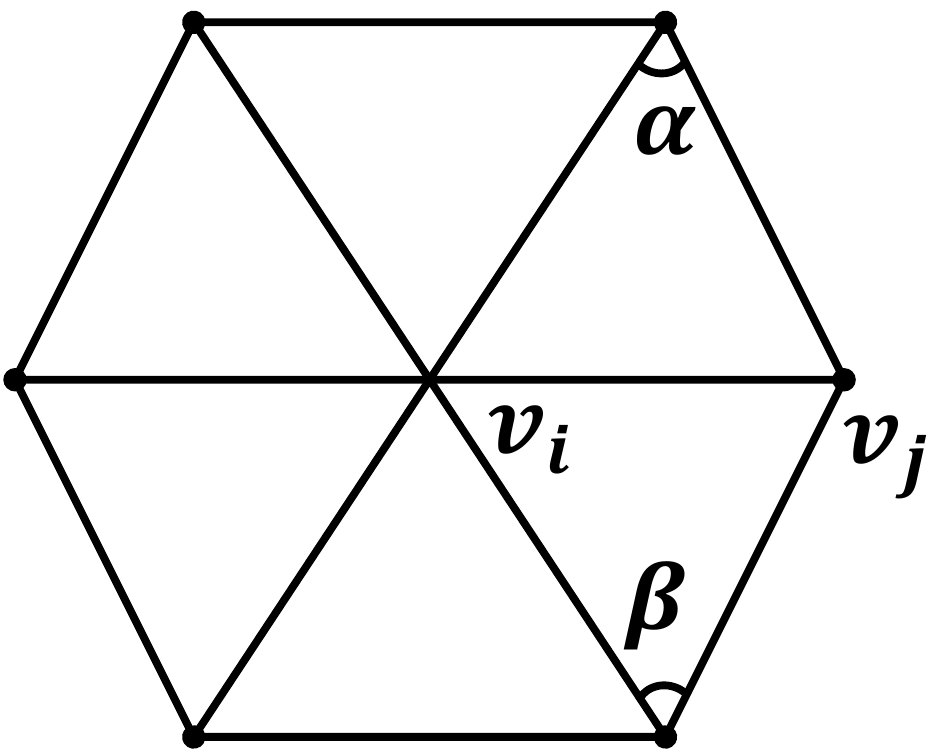}
\caption{Cotangent weight $w_{ij} = \dfrac{1}{2}(cot\alpha+cot\beta).$}
\label{fig:cotangent_weight}
\end{figure}

In the discrete setting, the above Poisson equation can be formulated as: for each vertex $v_i\in S$,
\begin{equation}
    \Delta f(v_i) := \sum_{v_i\sim v_j} w_{ij} (f(v_j)-f(v_i)) ,
    \label{eqn:discrete_Poisson}
\end{equation}
where $v_i\sim v_j$ means $v_j$ is adjacent to $v_i$, and  $w_{ij}$ is the well known cotangent weighting coefficients
\cite{gu2003global,duchamp1997hierarchical,desbrun2002intrinsic} as shown in Fig.~\ref{fig:cotangent_weight}.
Similarly, the co-boundary operator is
\begin{equation}
    \delta \omega(v_i) = \sum_{v_i\sim v_j} w_{ij}\omega([v_i,v_j]) .
    \label{eqn:coboundary}
\end{equation}
In our case, the cotangent weight is calculated under a special constant metric, see Fig.~\ref{fig:cotangent_weight}, all of the edges have an equal length $1$. It can be shown that the exact form extraction is independent of the choice of the metric. 
The exact component $df=(df_1,df_2)$ of $\omega=(\omega_1,\omega_2)$ can be obtained by directly solving the Poisson equation Eqn.~\ref{eqn:Poisson} using Eqn.~\ref{eqn:discrete_Poisson} and Eqn.~\ref{eqn:coboundary}.

%% file: experiments.tex
\section{Experimental Evaluations}
In this section, the experimental results are reported, which demonstrate the efficiency and accuracy of our global consistency optimization algorithm.
\begin{figure}[ht]
\centering
\includegraphics[width=1.0\textwidth]{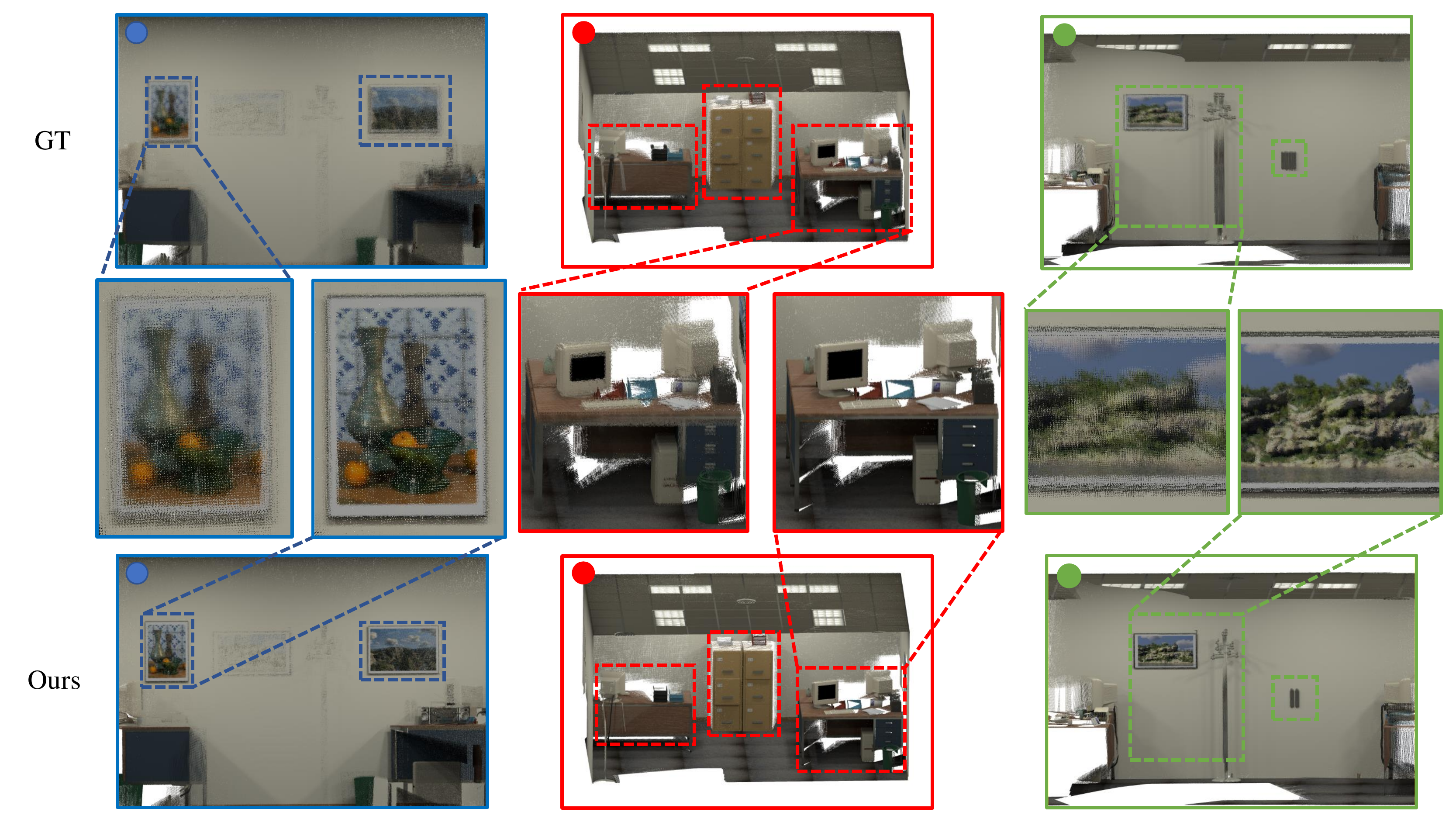}
\caption{Comparison between our method and the ``ground truth''(GT) of the Office 2 scene in the ICL-NUIM dataset. Top row: the reconstruction by the GT trajectory provided by the KinectFusion. Affected by the simulated sensor noises, the GT are imperfect. Bottom row: the registration result by our method. It can be seen that our method produces sharper details in the middle row. This demonstrates that our approach achieves higher quality.}
\label{fig.teaser}
\end{figure}
\begin{figure} 
\centering 
\includegraphics[width=1.0\linewidth]{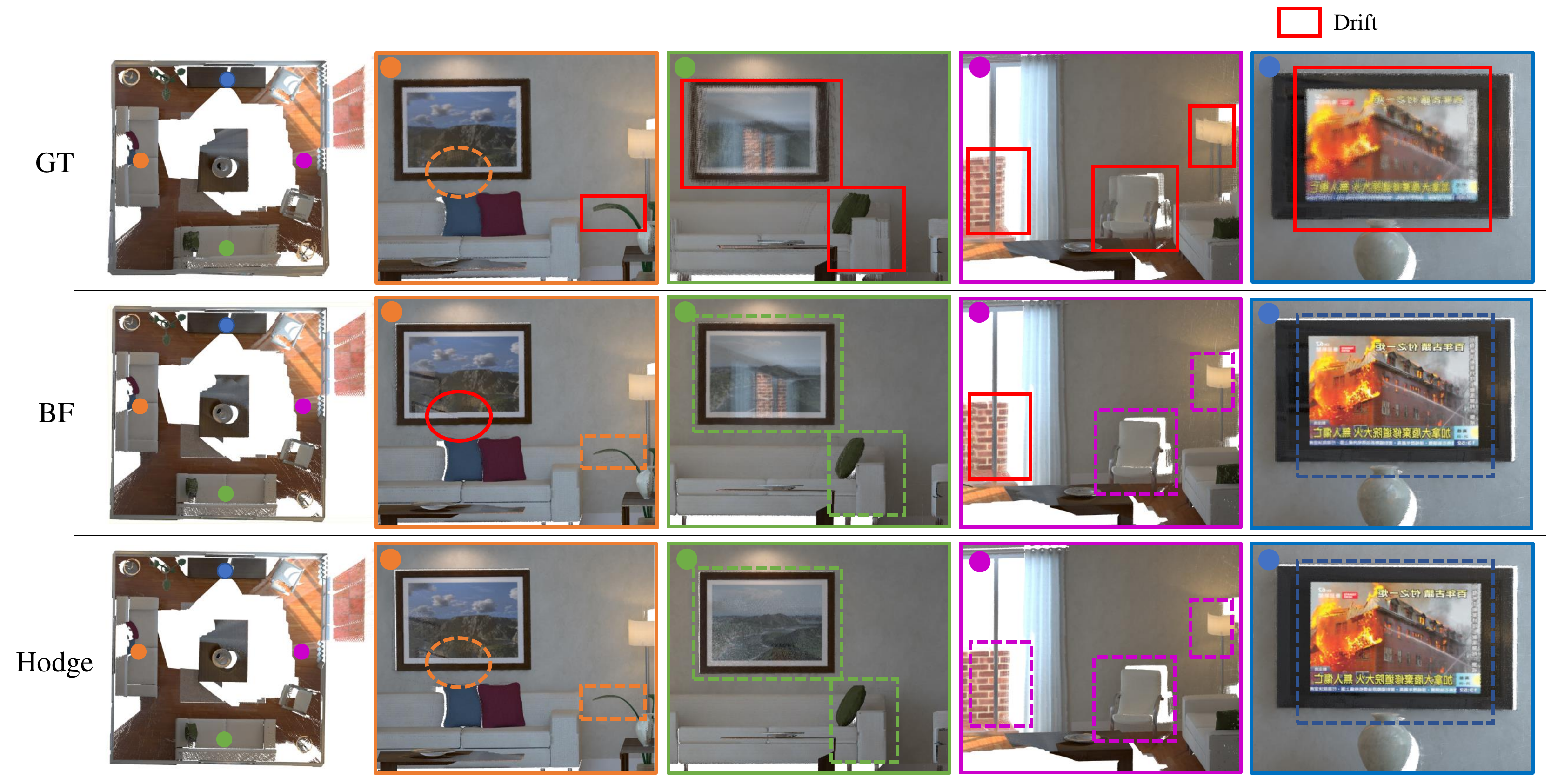}
\caption{Comparison among the ``ground truth'', the BundleFusion~\cite{BundleFusion} and our algorithm on the Living Room 2 scene in the ICL-NUIM dataset. Top row: the reconstruction (point cloud) by the ``ground true'' and the four local details (saffron:painting and plants, green:painting and sofa, purple:Balcony,armchair and lamp, blue:TV). Middle row: the reconstruction by the BundleFusion algorithm. Bottom row: the reconstruction by our method. The solid red box indicates poor reconstruction due to the  drifting effect, and the dotted box indicates the better results. It can be observed that our approach has the least drifts and highest global consistency.}
\label{icl-compare}
\end{figure}

\begin{figure}
  \centering
  \includegraphics[width=1.0\linewidth]{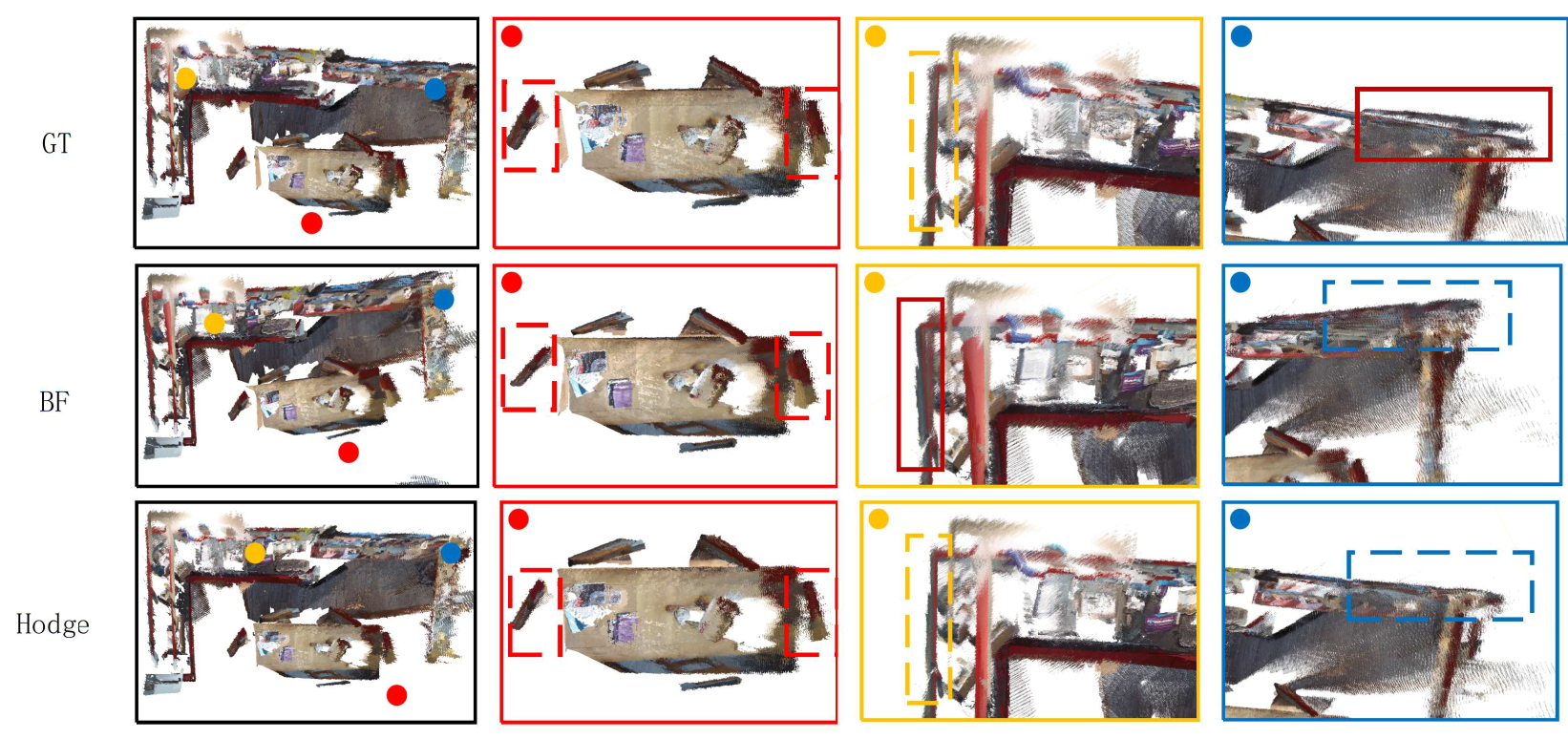}
  \caption{Comparison of the ``redkitchen" scene. Top row: the reconstruction (point cloud) by the ``ground truth'' and three local details (table, corner of upper left and right wall). Middle row: the  reconstruction by the Bundle-Fusion algorithm. Bottom row: the reconstruction by our method.The solid red curves mean the failure regions, the broken curves mean the regions without drifting. Numerically and visually, our method outperforms others. }
  \label{fig:redkitchen}
\end{figure}

\subsection{Datasets} 
The proposed algorithm has been evaluated on datasets from several different sources, including synthetic data with ground truth, real world data captured by handheld Kinect scanner and the geometric point clouds captured by our stereo-camera system with structured light.


\noindent{\textbf{ICL-NUIM Dataset}} consists of eight synthetic RGBD video sequences: four from an office scene and four from a living room scene, provided by Handa et al.~\cite{6907054}. 
The real camera pose information, obtained by the Kintinuous system~\cite{6631400}, allows us to accurately assess the accuracy of our reconstruction.
We conducted experiments on two input sequences, Living room 2 and Office 2, for a thorough comparison.\\

\noindent{\textbf{7-Scenes Dataset}}  released by Shotto et al~\cite{6619221}, consists of seven indoor scenes.
All seven scenes were shot with a handheld Kinect RGB-D camera and ``ground truth'' camera tracks were captured by the KinectFusion system~\cite{6162880}.\\

\subsection{Evaluation Procedure} 
\noindent{\textbf{Quantified Evaluation}} In order to quantitatively evaluate our method, we first construct the viewing graph based on two metrics, the centroidal distance and the intersection over union (IOU)~\cite{8886046}. We calculate the two metrics for each pair of point clouds. For a given pair, if the centroidal distance is less than a threshold or the IOU is greater than a threshold, then we add an edge corresponding to the pair of point clouds to the viewing graph.

For each edge in the viewing graph, we compute the \emph{negative exponential power of the point-2-plane distance}~\cite{8793516} between the two point clouds corresponding to the end nodes. The score is normalized to the unit interval, the higher score means the higher registration accuracy.



\noindent{\textbf{Experimental Results}} As explained in the algorithm section 4, our algorithm computes the initial Lie-algebraic 1-form $\omega$ using the ICP algorithm, then performs Helmholtz-Hodge decomposition by solving the Poisson Eqn.~\ref{eqn:Poisson} to extract the exact component $df$. Then $df$ is applied to update the poses and fuse the point clouds with high global consistency. If we want to further improve the global consistency and the reconstruction accuracy, we can repeat the whole procedure on the updated poses. Multiple iterations can be applied to achieve higher quality. 

For the comparison purpose, we transform the input point clouds (frames) along the available ``ground truth'' trajectory provided by the KinectFusion system or the Kintinuous system, and denote the results as the ``ground truth'' reconstructions. 

We compare the reconstruction results obtained by different algorithms, including the ``ground truth'' (GT), the BundleFusion (BF)~\cite{BundleFusion}, and our Hodge decomposition algorithm with one iteration (Hodge 1st) and three iterations (Hodge 3rd). The comparison results are summarized in Table ~\ref{scores on distance graph}  and Table ~\ref{scores on IoU graph}.

In Table ~\ref{scores on distance graph}, the viewing graph is constructed based on the centroidal distance among the point clouds, each edge corresponds to a pair of point clouds with centroidal distance less than $0.2$ meter. The scores of the negative exponential power of the point-2-plane distance on all the edges are averaged, the average scores for different algorithms are reported in the table. All the computations are with $6$ decimal precision, and the final results are rounded to reduce the redundancy. The best scores are in bold font. It can be seen that for $5$ of the $6$ examples, our proposed Hodge decomposition algorithm outperforms both the ground truth and the BundleFusion algorithm.

Similarly, in Table ~\ref{scores on IoU graph}, the viewing graph is constructed based on the intersection over union among the point clouds, each edge corresponds to a pair of point clouds with IOU greater than $0.5$. Our proposed method also outperforms the other two methods for $5$ of the $6$ examples. This demonstrates that our method can achieve higher reconstruction accuracy.

Note that in both cases, BundleFusion algorithm fails for reconstructing the office2 dataset, hence the pose for each frame can not be obtained. Hence its scores in both tables are labeled as nan.

\setlength{\tabcolsep}{4pt}
\begin{table}
\begin{center}
\caption{On the viewing graph under Euclidean metric,the average negative exponential power of the point-2-plane distance of ground turth, BundleFusion and Hodge method of iteration one and three time.The score's upper bound is 1, the larger the better.}
\label{scores on distance graph}
\begin{tabular}{lllll}
\hline\noalign{\smallskip}
 & \textbf{GT}      & \textbf{BF} & \textbf{Hodge 1st} & \textbf{Hodge 3rd} \\
\noalign{\smallskip}
\hline
\noalign{\smallskip}
   chess        & \textbf{0.829}   & 0.827       & 0.827           & 0.827           \\ 
   fire         & 0.869            & 0.868       & \textbf{0.870}  & 0.869           \\  
  redkitchen   & 0.863            & 0.865       & 0.865           & \textbf{0.865}  \\    
   stairs       & 0.821            & 0.818       & 0.822           & \textbf{0.822}  \\
   living2      & 0.803            & 0.883       & \textbf{0.921}  & 0.918           \\
  office2      & 0.783            & nan         & 0.931           & \textbf{0.937}  \\
\hline
\end{tabular}
\end{center}
\end{table}

\setlength{\tabcolsep}{4pt}
\begin{table}
\begin{center}
\caption{On the viewing graph under IoU metric, the average score of ground turth, BundleFusion and Hodge method of iteration one and three time. The score's upper bound is 1, the larger the better.}
\label{scores on IoU graph}
\begin{tabular}{lllll}
\hline\noalign{\smallskip}
 & \textbf{GT}      & \textbf{BF} & \textbf{Hodge 1st} & \textbf{Hodge 3rd} \\
\noalign{\smallskip}
\hline
\noalign{\smallskip}
   chess        & 0.837   & 0.835     & 0.836           & \textbf{0.838}           \\ 
   fire         & \textbf{0.870}   & 0.868     & 0.870           & 0.870           \\  
  redkitchen   & 0.862   & 0.864     & 0.864           & \textbf{0.865}  \\    
   stairs       & 0.818   & 0.814     & 0.820           & \textbf{0.820}  \\
   living2 & 0.805   & 0.878     & \textbf{0.919}  & 0.915           \\
   office2     & 0.781   & nan       & 0.930           & \textbf{0.935}  \\
\hline
\end{tabular}
\end{center}
\end{table}
\setlength{\tabcolsep}{1.4pt}
Furthermore, we also explicitly evaluate the global consistency by computing the integration of the Lie-algebraic 1-form along different loops as follows: first, we compute a set of the basis loops $\{\gamma_i\}_{i=1}^{2g}$ of the first homology group of the embedding surface $H_1(S,\mathbb{Z})$, and integrate $\omega$ along each base loop. Each integration result $\int_{\gamma_i}\omega$ is a rigid motion. We compute the $L^2$ norm of the translation component, and the Frobenius norm of the rotation component minus the identity matrix, the total norm measures the deviation of the rigid motion from the identity. 
We report the results below, which demonstrates the high global consistency (loop closedness) of our method.   

Fig.~\ref{fig.teaser} shows the comparison between our proposed algorithm and the ``ground truth'' on the ICL-NUIM dataset. It can be seen that the reconstruction of our method has higher quality and shows refiner local details.
Fig.~\ref{icl-compare} and Fig.~\ref{fig:redkitchen} shows the comparison among ground truth, BundleFusion~\cite{BundleFusion} and our algorithm on the ICL-NUIM dataset. By examining the reconstruction details, one can see that our method achieves the best quality and has the least drifting effects. 


\noindent{\textbf{Consistency Verification}}
We choose $m$ loops in the view graph, $\gamma_1,\gamma_2,\cdots,\gamma_m$. Suppose $\gamma_k$ has $n_k$ edges $\{e_{i,i+1}\}_{i=0}^{n_k-1}$, the rigid motion on the edge $e_{ij}$ is denoted as $M_{ij}^{k}$. The composition of the rigid motion along $\gamma_k$ is given by $\Pi_{i=0}^{n_k-1} M_{i,i+1}^k$. We measure the deviation of the composed rigid motions from the identity for all the loops, and denoted it as
\begin{equation}
D:=\sum_{i=1}^m \|  \Pi_{i=0}^{n_k-1} M_{i,i+1}^k - I_4 \|_F.  
\label{eqn:loop closedness}
\end{equation}
where $I_4$ is the fourth order identity matrix. Deviation $D=0$ means that the relative poses are consistent along all loops, i.e. they satisfies the global consistency condition: loop closedness.

For the ICL-NUIM and 7-Scenes dataset, we list all the deviations $D$'s in Eqn.~\ref{eqn:loop closedness} of three iterations of our Hodge decompose algorithm,  denoted as Hodge 1, Hodge 2 and Hodge 3 respectively in Table~\ref{Loop consistency on ICL-NUIM and 7-Scenes dataset}.

In Table~\ref{Loop consistency on ICL-NUIM and 7-Scenes dataset}, it can be seen that among the first $6$ examples, the Hodge decomposition with one iteration obtains relative large deviations, by increasing the iterations, the global consistency is greatly improved, 
two or three iterations can achieve zero deviations $D$'s.

\setlength{\tabcolsep}{4pt}
\begin{table}
\begin{center}
\caption{Loop consistency on ICL-NUIM and 7-Scenes dataset} 
\label{Loop consistency on ICL-NUIM and 7-Scenes dataset} 
\begin{tabular}{llllllll}
\hline\noalign{\smallskip}
     & living2 & office0  & office2  & office3  & redkitchen  & chess & other\\
\hline   \textbf{Hodge 1}    & 3e-4  & 4.16e-2   & 3e-4   & 1.0463e-1  & 7.28e-2   & 1.5e-4 & 0.00\\
\hline   \textbf{Hodge 2}      & 0.00    & 2e-5  & 0.00     & 0.00     & 0.00       & 0.00 & 0.00\\
\hline   \textbf{Hodge 3}      & 0.00     & 0.00      & 0.00     & 0.00      & 0.00        & 0.00 & 0.00\\
\hline
\end{tabular}   
\end{center}   
\end{table}



%% file: conclusion.tex
\section{Conclusions}
This work introduces a novel algorithm to tackle the challenging loop closing problem in global point cloud registration. The algorithm generalizes the viewing graph to a Cech complex with richer topological structure, and treat the relative transformations defined on the edges as a Lie-algebraic simplicial 1-form on the Cech complex, then use Helmholtz-Hodge decomposition to extract the exact component of the 1-form, which gives the relative transformations among the frames. Furthermore, the 2-skeleton of the Cech complex can be approximated by an embedding surface of the viewing graph.

Experimental results demonstrate the simplicity, efficiency and accuracy of the algorithm. It outperforms the conventional methods and achieves higher global consistency.

However, this algorithm assumes the relative transformations on each edge in the viewing graph is in the neighborhood of the identity in the Lie group of the rigid motion, hence it can be represented using a vector in the Lie algebra. If the poses are sparse, and the relative transformations are far away for the identity, the Lie-algebraic representations are inaccurate. The proposed method will encounter difficulties. In the future, we will explore further to overcome this shorting coming.